\documentclass[conference]{IEEEtran}
\IEEEoverridecommandlockouts

\usepackage[T1]{fontenc}
\usepackage{graphicx}
\usepackage{booktabs}
\usepackage[misc]{ifsym}
\usepackage{multirow}
\usepackage{subcaption}
\usepackage{caption}
\usepackage{amsmath}
\usepackage{times} 
\usepackage{lipsum} 
\usepackage{algorithm}
\usepackage{algpseudocode}
\usepackage{tikz,pgfplots,subcaption}
\usepackage{hyperref}
\usepackage{booktabs} 
\usepackage{siunitx}  
\usepackage{colortbl}
\usepackage{hhline}
\usepackage{enumitem}
 \usepackage{etex} 
\usepackage{mwe}
\usepgfplotslibrary{external} 
\def\BibTeX{{\rm B\kern-.05em{\sc i\kern-.025em b}\kern-.08em
    T\kern-.1667em\lower.7ex\hbox{E}\kern-.125emX}}
\begin{document}

\title{Calibrating Practical Privacy Risks for Differentially Private Machine Learning\\
\thanks{This work is partially supported by the National Science Foundation (Aware \# 2232824). }
}

\author{\IEEEauthorblockN{1\textsuperscript{st} Yuechun Gu}
\IEEEauthorblockA{\textit{TAIC Lab} \\
\textit{UMBC}\\
Baltimore, USA \\
ygu2@umbc.edu}
\and
\IEEEauthorblockN{2\textsuperscript{nd} Keke Chen}
\IEEEauthorblockA{\textit{TAIC Lab} \\
\textit{UMBC}\\
Baltimore, USA \\
kekechen@umbc.edu}
}

\maketitle

\begin{abstract}
Differential privacy quantifies privacy through the privacy budget $\epsilon$, yet its practical interpretation is complicated by variations across models and datasets. Recent research on differentially private machine learning and membership inference has highlighted that with the same theoretical $\epsilon$ setting, the likelihood-ratio-based membership inference (LiRA) attacking success rate (ASR)  may vary according to specific datasets and models, which might be a better indicator for evaluating real-world privacy risks. Inspired by this practical privacy measure, we study the approaches that can lower the attacking success rate to allow for more flexible privacy budget settings in model training. We find that by selectively suppressing privacy-sensitive features, we can achieve lower ASR values without compromising application-specific data utility. We use the SHAP and LIME model explainer to evaluate feature sensitivities and develop feature-masking strategies. Our findings demonstrate that the LiRA $ASR^M$ on model $M$ can properly indicate the inherent privacy risk of a dataset for modeling, and it's possible to modify datasets to enable the use of larger theoretical $\epsilon$ settings to achieve equivalent practical privacy protection. We have conducted extensive experiments to show the inherent link between ASR and the dataset's privacy risk. By carefully selecting features to mask, we can preserve more data utility with equivalent practical privacy protection and relaxed $\epsilon$ settings. The implementation details are shared online at the provided GitHub URL \url{https://anonymous.4open.science/r/On-sensitive-features-and-empirical-epsilon-lower-bounds-BF67/}. 
\end{abstract}

\begin{IEEEkeywords}
Differential privacy, Feature sensitivity, Membership inference attack
\end{IEEEkeywords}

\section{Introduction}
With the fast advancement of deep learning techniques, companies can now leverage a huge amount of user-generated or user-related image and text data to train powerful large models. A main concern is these large models learn much more than what they are supposed to learn \cite{Zytek22} -- once the models are published, adversaries can use them to infer private information in the training data, e.g., via model inversion \cite{fredrikson2015model,zhang2020secret}, membership inference \cite{shokri2017membership,choquette2021label}, property inference \cite{ganju2018property,parisot2021property}, and domain inference \cite{gu2023gan,gu2023adaptive} attacks. To address the private-information leak from published models, recent developments in privacy-preserving deep learning have incorporated the theory of differential privacy \cite{gong2020survey}, e.g., the well-known Differentially Private Stochastic Gradient Descent (DP-SGD) \cite{abadi2016deep}.

A unique feature of differentially private machine learning is that the setting of privacy budget $\epsilon$ in $(\epsilon, \delta)$-differential privacy \cite{dwork2008differential} is independent of applications and data distributions. The smaller the $\epsilon$ setting, the better the privacy is preserved. It's considered an advantage since the learned privacy-preserving model will be resilient to attacks equipped with any type of adversarial knowledge. On the other side, it’s also well believed that the differential privacy setting is too conservative to preserve data utility \cite{jorgensen2015conservative,el2022differential}. For instance, the commonly accepted setting $\epsilon=1$ has led to significant utility loss for many applications. In real-world applications, much higher $\epsilon$ values are often used to achieve better data utility, which has raised concerns among researchers \cite{Apple,aktay2020google} since no clear guidance is available to justify this practice.  An intriguing question is whether and how we can relax the $\epsilon$ setting for different types of data and applications\footnote{Intuition says yes to the first question: When it comes to tasks that do not contain personally identifiable information, e.g., classifying animal pictures, we may use an arbitrary large $\epsilon$.}.

Is there a more practical auxiliary measure that can guide us to set the privacy parameters?  This measure is likely dataset- and model-specific to complement the application-agnostic nature of differential privacy. The recent development on membership inference attacks (MIA) reminds us that it’s possible. However, nobody has tried to apply MIA to this challenging problem.

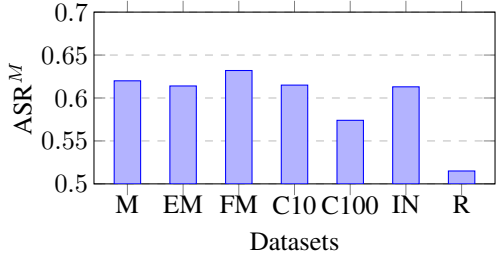
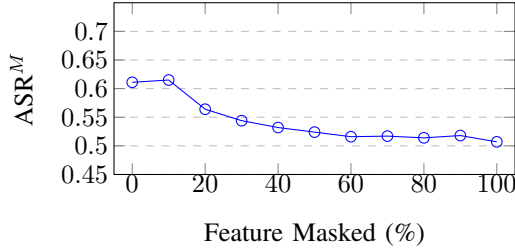
\begin{figure}[htbp]
  \centering
  \begin{subfigure}[t]{0.42\textwidth}
    \centering
    \begin{tikzpicture}
      \begin{axis}[
          ybar,
          scale only axis,
          width=0.7\textwidth,
          height=0.3\textwidth,
          ymajorgrids=true,
          grid style=dashed,
          xlabel = {Datasets},
          ylabel near ticks,
          ylabel={ASR$^M$},
          symbolic x coords={M,EM,FM,C10,C100,IN,R},
          xtick=data,
          ytick={0.5, 0.55, 0.6, 0.65, 0.7},
          ymin=0.5,ymax=0.7,
          x tick label style={rotate=0, inner sep=0pt, xshift=0pt},
          y tick label style={rotate=0, inner sep=0pt, xshift=-2pt}
      ]
        \addplot coordinates {
          (M,0.62) 
          (EM,0.614) 
          (FM,0.632) 
          (C10,0.615) 
          (C100,0.574)
          (IN,0.613)
          (R,0.515)
        };
      \end{axis}
    \end{tikzpicture}
    \caption{$ASR^M$ variation across datasets. The model trained on a randomly generated dataset has $ASR^M \approx 0.5$.}
    \label{f1a}
  \end{subfigure}
  \hfill
   \begin{subfigure}[t]{0.42\textwidth}
    \centering
    \begin{tikzpicture}
      \begin{axis}[
          xlabel={Feature Masked (\%)},
          width=0.7\textwidth,
          height=0.3\textwidth,
          xmin=-5, xmax=105,
          ymin=0.45, ymax=0.75,
          xtick={0,20,40,60,80,100},
          ytick={0.45,0.5, 0.55, 0.6, 0.65,0.7},
          legend pos=north east,
          ymajorgrids=true,
          grid style=dashed,
          scale only axis,
          grid style=dashed,
          ylabel={ASR$^M$},
          x tick label style={rotate=0, inner sep=0pt, xshift=0pt},
          y tick label style={rotate=0, inner sep=0pt, xshift=-2pt}
      ]
        \addplot[color=blue, mark=o] coordinates {
          (0,0.611) 
          (10,0.615) 
          (20,0.564) 
          (30,0.544) 
          (40,0.532) 
          (50,0.524) 
          (60,0.516) 
          (70,0.517) 
          (80,0.514) 
          (90,0.518) 
          (100,0.507)
        };
      \end{axis}
    \end{tikzpicture}
    \caption{$ASR^M$ decreases with more masked features (MNIST).}
    \label{f1c}
  \end{subfigure}
  \caption{$ASR^M$ variation across different datasets and modified versions of the same dataset. All models are trained using DP-SGD with a theoretical $\epsilon=8$. Abbreviations: M (MNIST), EM (EMNIST), FM (Fashion-MNIST), C10 (CIFAR-10), C100 (CIFAR-100), IN (ImageNet-1k subset), R (Randomly generated dataset).}
  \label{fig:performance_epsilon}
\end{figure}

Carlini et al. \cite{carlini2022membership} show that the likelihood- ratio-based membership inference attack (LiRA) can utilize the definition of differential privacy directly to conduct a sample-level membership inference attack. When applied to machine learning models, the essential idea of differential privacy is interpreted as follows: it’s difficult (at the $\exp^\epsilon$ level) to distinguish whether a model used a sample in training or not. LiRA conducts a hypothesis-testing approach to derive the likelihood of a sample’s membership. Based on LiRA, one can also derive the attacking success rate (ASR) for each sample \cite{carlini2022onion}. Due to the intrinsic link between LiRA and differential privacy \cite{Ahmed23}, it’s possible to use LiRA-based ASR as a measure to guide the selection of $\epsilon$. We define a dataset-level $ASR^M$ to indicate the worst-case identifiability of all samples in terms of a model $M$. We find this measure can precisely capture the \emph{sensitivity level of dataset} in machine learning. Specifically, $ASR^M = 0.5$ indicates that LiRA is equivalent to random guessing. Thus, all samples are safe, the dataset is not sensitive to the model $M$, and we can use a larger $\epsilon$ setting. $ASR^M\approx 1$ indicates the LiRA can correctly identify almost every participating sample. Thus, the dataset is highly sensitive under the model $M$, and a smaller $\epsilon$ value is needed to protect sample privacy. Initial evidence shows that this measure is indeed dataset- and model-specific. Figure~\ref{f1a} shows that $ASR^M$ varies significantly over different datasets and corresponding models and correctly shows $\approx 0.5$ for randomly generated datasets. 

More interestingly, since LiRA and $ASR^M$ are dataset-specific, we can modify the dataset to lower its sensitivity level. 

A simple experiment shows that $ASR^M$ decreases with the increasing number of randomly masked features (see Figure \ref{f1c}). This gives us an opportunity that \emph{we may adjust the sensitivity level of a dataset to relax the $\epsilon$ setting in DP-SGD.} Such a method is also possibly optimized to achieve good utility and privacy preservation.

\textbf{Scope of Research.} This paper studies two important problems: (1) a practical method for guiding the selection of $\epsilon$ value in differentially private machine learning, and (2) modifying a dataset to allow more relaxed $\epsilon$ settings to achieve a better balance between privacy and utility. 

We have adopted the attacking success rate (ASR) of LiRA as the basic measure to indicate the practical privacy risk of a dataset under a specific model release, due to its tight connection with the definition of differential privacy. We have analyzed why this measure is data- and model-specific based on the definition of the hypothesis testing method (Section \ref{ft_masking}).

We further investigate possible dataset modification strategies that can lower the sensitivity of the dataset while not significantly damaging its utility. With lowered dataset sensitivity, we can apply relaxed $\epsilon$ settings, which can help achieve much better utility. The basic dataset-sensitivity reduction strategy is inspired by feature suppression used by data anonymization \cite{fung10}, i.e., masking sensitive features to protect privacy. Our method incorporates model explanation techniques, such as LIME \cite{ribeiro2016why} and SHAP \cite{lundberg2017unified}, to identify \emph{utility-sensitive} and \emph{privacy-sensitive} features, respectively, from the target utility task and an auxiliary identity-related task. For example, distracted driver classification is an important utility task for smart vehicles, which helps identify tired or distracted drivers and prevent potential car accidents. The training data contains (input: driver image, label: type of distraction) pairs. It’s easy to define an auxiliary identity task by replacing the labels with the identities of the drivers. We find that if the top features (e.g., pixels in images) from both types of tasks are not entirely overlapping, we can always extract a subset of features to suppress that helps reduce the dataset sensitivity while minimizing the loss of data utility. The lowered dataset sensitivity allows us to set a higher $\epsilon$ value in DP-SGD. 

We show that with our approach, we can achieve the same level of practical privacy protection (i.e., ASR$^M$) with much better-preserved utility in experiments. We have used well-known datasets, e.g., MNIST, CIFAR10, facial expression datasets, and distracted driver detection datasets \cite{100Driver} in experiments. Facial expression and distracted driver detection datasets are adopted to intuitively show how the identity and utility tasks are defined in our approach. On feature-masked datasets, we observe low ASR$^M$ values (the practical privacy threats) even with theoretical $\epsilon$ values 5-10 times larger than the unmasked ones and 22\%-41\% better model quality. 

In summary, we conclude our contributions as follows:

\begin{itemize}[noitemsep]
    \item We are the first team to investigate the data and model-specific nature of the LiRA attack, which can guide the selection of theoretical $\epsilon$ settings for specific modeling tasks.
    \item We show that the LiRA-based attacking success rate can be reduced via masking features of the target dataset, which can be optimized to achieve a better balance between utility and privacy.
    \item We have demonstrated that our methods work as expected in experiments on real datasets and modeling tasks. We have also shared the source code for researchers to reproduce the results.
\end{itemize}

The remaining sections cover the following topics. We explore related works in Section \ref{sec:related}, introduce key background information in Section \ref{sec:prelimineries}, provide insights into the core ideas of this paper, and how we put our feature masking method into action in Section \ref{ft_masking}. Results from our experiments are discussed in Section \ref{sec:exps}. Lastly, we conclude our work in Section \ref{sec:conclusion}.

\section{Related Works}
\label{sec:related}
Differential privacy, established by Dwork et al. \cite{dwork2006differential}, is the most well-accepted theoretically justifiable privacy protection method. It has been integrated into deep learning via methods like DP-SGD \cite{abadi2016deep}, DP-Adam, and DP-RMSProp \cite{dp-adam}. Despite its strong theoretical guarantees, the application-specific setting of the privacy budget ($\epsilon$) remains elusive, as the theoretical privacy budget $\epsilon$ setting is agnostic to datasets and models \cite{dwork2014algorithmic}. Real-world applications often use a relaxed setting, e.g., Apple's reported $\epsilon$ values of 2 to 16 for user activity analysis \cite{Apple} and Google's 2.64 for COVID-19 community mobility reports \cite{aktay2020google}, which raise concerns about the actual protection power.

Carlini et al. \cite{carlini2022membership} proposed a likelihood ratio attack (LiRA) based on a black-box membership inference game using hypothesis testing. In their offline version of LiRA, adversaries aim to calculate the probability of rejecting the hypothesis that a target sample is not part of the training data distribution. According to Ahmed et al. \cite{Ahmed23}, since LiRA operates within a black-box membership inference game, there exists a differential privacy distinguisher that gives the same results as  LiRA’s.

Some researchers also combine differential privacy with feature selection. Zhang et al. \cite{Zhang20} consider the issue of privacy loss when data have a correlation in machine learning tasks and use differential privacy to privately select important features from datasets to avoid compromising privacy in the correlation of features. Pittaluga et al. \cite{Pittaluga2023_ICCV} use differential privacy to privatize the features in the sample and design a feature-level differential privacy to guarantee privacy. However, none of the existing works interpret and guide the $\epsilon$ setting from the perspective of feature importance

Feature suppression was used in data anonymization \cite{fung10} to hide those features that generalization does not work. We use it to lower a given dataset's sensitivity level, i.e., the LiRA attacking success rate $ASR^M$. It explores a possible way of combining traditional data anonymization methods and differential privacy to achieve better utility preservation with solid theoretical privacy guarantees.

\section{Prelimineries}
\label{sec:prelimineries}
\subsection{Differential Privacy}
\label{sec:DP_Auditing}
Differential Privacy (DP) aims to enable the analysis of a dataset while ensuring that the inclusion or exclusion of any single individual's data does not significantly affect the outcome of the analysis. Mathematically, DP is defined as follows: 

\textbf{($\epsilon, \delta$)-Differential Privacy (DP).} Let $\mathcal{A}$ be a randomized function. We say that $\mathcal{A}$ provides $(\epsilon, \delta)$-differential privacy if for all datasets $D_0$ and $D_1$ differing on at most one element, and for all $O \subseteq \text{Range}(\mathcal{A})$,
$$\Pr[\mathcal{A}(D_0) \in O] \leq e^\epsilon \times \Pr[\mathcal{A}(D_1) \in O] +\delta$$
where $\delta$ is an ignorable small value, often $\ll N$, the number of samples in the dataset, and $\epsilon$ is often called the privacy budget. In the context of machine learning, the adversary's ability to distinguish if the model $\mathcal{A}$ is trained on $D_0$ or $D_1$ is bounded by $e^\epsilon$, and $\delta$ is the probability that the bound fails to hold. 

Since a differentially private deep learning algorithm, such as DP-SGD \cite{abadi2016deep}, involves many steps of calculation and randomization, a specific privacy budget accounting method is used to aggregate step-wise privacy budgets to derive the estimate of the overall privacy budget $\epsilon$. This estimate is often called the \emph{theoretical $\epsilon$}, which is often considered conservative, i.e., the actual $\epsilon$ is smaller.  

\textbf{Offline Likelihood-ratio test.} Following the definition of differential privacy where $D$ and $D/x$ differ in sample x, Carlini et al. \cite{carlini2022membership} present both the online and offline LiRA hypothesis testing methods. The online LiRA test requires attackers to train multiple shadow models on both in-domain and out-domain datasets, resulting in extremely high computational costs. In contrast, the offline LiRA test does not use in-domain information and relies solely on out-domain samples, which significantly reduces computational costs. Therefore, we choose the attacking success rate ($ASR$) of the offline LiRA test as the indicator of the sensitivity of sample $(x,y)$.
To test the membership of $(x,y)$, we measure the probability of observing confidence as high as the target model's under the null hypothesis that the target point $(x,y)$ is a non-member as follows:

\begin{align*}
\Lambda((x,y)) &= 1 - \Pr[Z > \phi(q)], \text{  where}\quad Z \sim \mathcal{N}(\mu_{\text{out}}, \sigma_{\text{out}}^2) \\
   \phi(q) &= \log \frac{q}{1-q}, \text{  where} \quad q = M(x)_y
\end{align*}

For a new sample $x$ to be tested, we apply the specific model $M(x)$ and test whether its output's log-likelihood-ratio $\log\frac{q}{1-q}$ is significantly higher than the typical out-domain's, i.e., we determine the new sample as a member sample when $\Lambda(x)>$ threshold. The threshold is set to 0.5 \cite{carlini2022membership}.

Based on this test, we can estimate the $ASR$ of each sample in the dataset in a batch-based method. Specifically, we start by training $n$ shadow models $M_{0\dots n}$ on $n$ randomly sampled subsets of the original dataset, which serve as the hypothetical in-domain shadow datasets. Each shadow dataset, $D_{IN}^i$, is constructed by randomly picking samples from the entire dataset a probability of 0.5, and the remaining unselected samples are the out-domain samples, $D_{OUT}^i$. $D_{OUT}^i$ samples go through the model $M_i$ and the log-likelihood conversion of their outputs roughly follows a normal distribution, which is used to estimate the distribution parameters $\mu_{out}$ and $\sigma_{out}^2$. For each sample $x_j$ in the dataset, we have $n$ attacking results $\text{LiRA}(x_j,M_i)$ which will be used to compare to the ground truth label $G(x_j,M_i) = \text{``in-domain’’ or ``out-domain’’}$. We compute the $ASR$ of sample $x_j$ as follows:
$$ ASR(x_j) = \sum_{i=1}^n \mathbf{1}(\text{LiRA}(t_j, M_i) == G(t_j, M_i)) / n $$

\textbf{Dataset Sensitivity.} Furthermore, following the worst-case scenario of differential privacy, we can also measure the overall sensitivity of a dataset under a model $M$ with $ASR^M = \max ASR(x_j), j=1..N$, for $N$ samples in the dataset. Intuitively, the highest sample ASR determines the sensitivity level of the dataset, i.e., the worst case of the dataset under the attack.  

\section{Reducing Dataset Sensitivity via Feature Masking}
\label{ft_masking}
In this section, we begin by discussing the rationale behind our method and outlining our threat modeling. Then, we present the main concepts and definitions of feature masking for reducing the empirical lower bound.

\subsection{Motivation and Threat Modeling}
\label{sec:threatmodeling}
Differentially private machine learning aims to train machine learning or deep learning models that are resilient to privacy attacks, such as membership inference \cite{hu2022membership}. A common DP learning algorithm, e.g., DP-SGD, allows the model owner to specify the privacy budget $\epsilon$ in $(\epsilon, \delta)$-differential privacy -- the smaller the $\epsilon$, the better the privacy is preserved. However, this privacy setting is independent of the dataset and trained model, ignoring practical privacy risks the model may have. Our purpose is to study empirical privacy risk to derive better data- and model-specific privacy measures. 
The initial results on the LiRA (Section \ref{sec:DP_Auditing}) imply that the privacy level of one sample can be estimated with the hypothesis testing version of the membership inference attack. 
Before diving into the details, we introduce the threat modeling for privacy attacks under the deep learning environment.

\textbf{Protected assets:} Identities of training data examples, i.e., the user who contributed or is related to a sensitive training sample. Examples: samples can be used to directly identify the owner, e.g., a face image; or samples contain unique features that can be used to link other data sources, e.g., tooth images, if linked to a person's dental records, can be used to identify a person.

\textbf{Involved parties:} Data and model owners and attackers. Data and model owners may use differential privacy to disguise the training of the model. The training process is secure and private, and no information is leaked. The learned model might be exposed via a service interface. 

\textbf{Adversarial capability.} We assume the learned model might be exposed via a service interface that might be under either white-box or black-box privacy attacks. A white-box attack can access the internal structure of the model, while a black-box attack can only use the model prediction service. We also assume attackers know what the model is used for and the training data distribution but not individual training data examples. Attackers will try to breach the privacy of individual training data examples, e.g., via membership inference attacks.

\subsection{LiRA Attacking Success Rate is Data and Model Specific} 
We examine the definition of the offline version of LiRA to show that the $ASR$ is data and model-specific.  
Note the hypothesis testing method needs to train models, $M_1\dots M_n$, each of which holds $H_0$ (does not contain the canary examples) true or not. Each test is done by determining whether $\Lambda((x,y))>$ threshold holds as decipted in Section~\ref{sec:prelimineries}. Therefore, the testing is inherently tied to the model and the dataset. 

On random datasets, all the models essentially perform random guessing, and $ASR^M \approx 0.5$. As expected, we have found that $ASR^M$ varies by datasets and models in experiments. Furthermore, we have noticed by modifying the dataset, e.g., randomizing the labels or removing sensitive features, $ASR$ can be significantly reduced (Figure~\ref{fig:performance_epsilon}). 
This observation inspired us to find ways to reduce $ASR^M$ for a specific dataset. We consider a few candidate methods as follows. 

\emph{1. Lowering $\epsilon$.} In addition to the inherent randomness of the dataset, a lower privacy budget $\epsilon$ may push down the empirical privacy risk as well, but it also reduces more data utility, which is against our goal. However, it's still important to learn how the measure $ASR^M$  changes with the setting of  $\epsilon$. If they are positively correlated, we can use this correlation to guide the setting of the privacy budget for a modified dataset, as we will show in the experiments.

\emph{2. Reducing dataset sensitivity.} Since differential privacy is designed to protect the privacy of human-related data, intuitively, each dataset is associated with some level of \emph{privacy sensitivity}, depending on how it can be used to explore private information. There are candidate methods for lowering dataset sensitivity. (1) Injecting noise at the instance level to disguise the private information and thus reduce the overall dataset sensitivity. This approach is adopted by locally differential privacy \cite{yang2023local}, which, however, leads to significantly more utility loss than the global differential privacy approach used by DP-SGD. (2) Identifying and removing privacy-sensitive features (i.e., the feature masking approach). If we model a privacy attack as a learning task -- recovering the human-related identity information from the training examples, we might be able to identify the features that help this task most and then remove them. Experiments show that such a procedure indeed can help reduce dataset sensitivity represented with $ASR^M$.

\textbf{Approximately equivalent $\epsilon$ settings.} A key hypothesis is whether different $\epsilon$ settings for the original and modified datasets provide equivalent \emph{practical} privacy protection. So far, there is no method for finding the exact practical privacy guarantee for a theoretical $\epsilon$ setting. However, we think the attacking success rate of LiRA might be a close one indicating a practical privacy guarantee. We thus define the equivalency as follows. 

Let $M_{\epsilon, D}$ be a model trained with $(\epsilon, \delta)$ differential privacy on $D$, and $M_{\epsilon, D'}$ trained with $(\epsilon', \delta)$ differential privacy on a modified dataset of $D$. The ASRs are $ASR^{M_{\epsilon, D}}$ and $ASR^{M_{\epsilon', D'}}$, correspondingly. Assuming a small $\delta$ (e.g., $\delta = 10^{-5}$) is ignorable. We say the settings $\epsilon$ for $D$ and $\epsilon'$ for $D'$ are approximately equivalent, if$$ASR^{M_{\epsilon, D}} \approx ASR^{M_{\epsilon', D'}}$$

In the following sections, we will explore the idea of feature masking for achieving lower $ASR$ and then optimize the masking methods to preserve utility as well.

\subsection{Optimized Feature Masking}
\label{sec:optimization}
\textbf{Identity and utility tasks.} Instead of examining a real privacy attack on the target model or dataset to identify the feature sensitivity level, we use \emph{an identity task} as the surrogate to understand the feature sensitivity of the dataset. In contrast, we name the original modeling task as \emph{the utility task}. It's best to understand the two tasks in terms of real applications. For example, a facial expression recognition task is a utility task. Since the training data is collected from several persons' expressions, the data can also be used to learn who might be the contributors, which is defined as the identity task. In another example, a self-driving dataset is originally used to identify all kinds of objects from the captured scene: roads, trees, sidewalks, pedestrians, etc., which is the utility task. The identity task can be whether the scene contains persons, which can be used to rank the person-related features that are potentially linked to privacy protection.  

One may argue that such an identity task may not be easy to define in some applications. The role of identify task represents the model builder’s \emph{best} knowledge about privacy sensitive information. It provides some hints to our approach that masking certain features might reduce the ASR$^M$ value. With the ASR$^M$ calculation procedure, the model builder can always verify whether the masking helps privacy protection and whether it damages utility.  

\textbf{Feature privacy sensitivity.} We consider a sensitive dataset $D$ comprising $N$ samples $\{x_i\}, i=1..N$, each with $K$ features $\{F_j\}, j=1..K$.  The first crucial step in our method is to ascertain the sensitivity level of each feature in $D$. This sensitivity level, which we term the \emph{Feature Privacy Sensitivity} ($s_j$) for each feature $F_j$, is a pivotal factor in our feature masking strategy.
With a surrogate identity model $I(D, Z): z = g(x), z \in Z$, where $Z$ is a set of identities, we design a method to quantify the feature-level sensitivity. Intuitively, this model, trained on a labeled dataset $\{(x_i, z_i)\}$, with crafted identity-related labels $z_i \in Z$, can output person-related information, e.g., whether an image $x$ contains a person $z$. The feature importance, as it tells the feature's contribution to the identity task, can be used to define the feature privacy sensitivity $s_j$, which can be captured by a model explainer, like SHAP \cite{lundberg2017unified} or LIME \cite{ribeiro2016why}. The effect of different explainers has been evaluated in experiments. For simplicity, assume we use SHAP values of the identity task as the feature sensitivity. We only consider positive SHAP values as they indicate the features' positive contribution to the identity task:
\[
s_j = 
\begin{cases}
    s_j & \text{if } s_j > 0 \\
    0 & \text{otherwise.}
\end{cases}
\]

Upon determining the feature privacy sensitivity, our next step is to mask features based on their sensitivity levels and their values to the utility task. We consider both privacy loss and data utility and try to achieve an optimized balance between the two as follows

\textbf{Utility-optimized masking.}  Similar to the definition of feature privacy sensitivity, e.g., with SHAP values, we can also derive feature utility sensitivity in terms of the utility task and a utility model $M(D,Y)$, where $Y$ is a set of utility labels, e.g., facial expressions in facial expression datasets. We define the feature $F_j$'s \emph{utility sensitivity}, $u_j$, as follows:
\[
u_j = 
\begin{cases}
    u_j, & \text{if } u_j > 0 \\
    0, & \text{otherwise}
\end{cases}
\]
Let the dot-product of the normalized utility-sensitivity vector $u$ and privacy-sensitivity vector $s$ represent the extent of shared features between the identification and utility models. Ideally, when $u$ and $s$ are orthogonal, i.e., $u^Ts = 0$, there is no overlap between identity-related and utility-related features. This would make it straightforward for the data owner to eliminate identity-related features without affecting utility.

\begin{figure}[htbp] 
   \centering
   \includegraphics[width=0.7\linewidth]{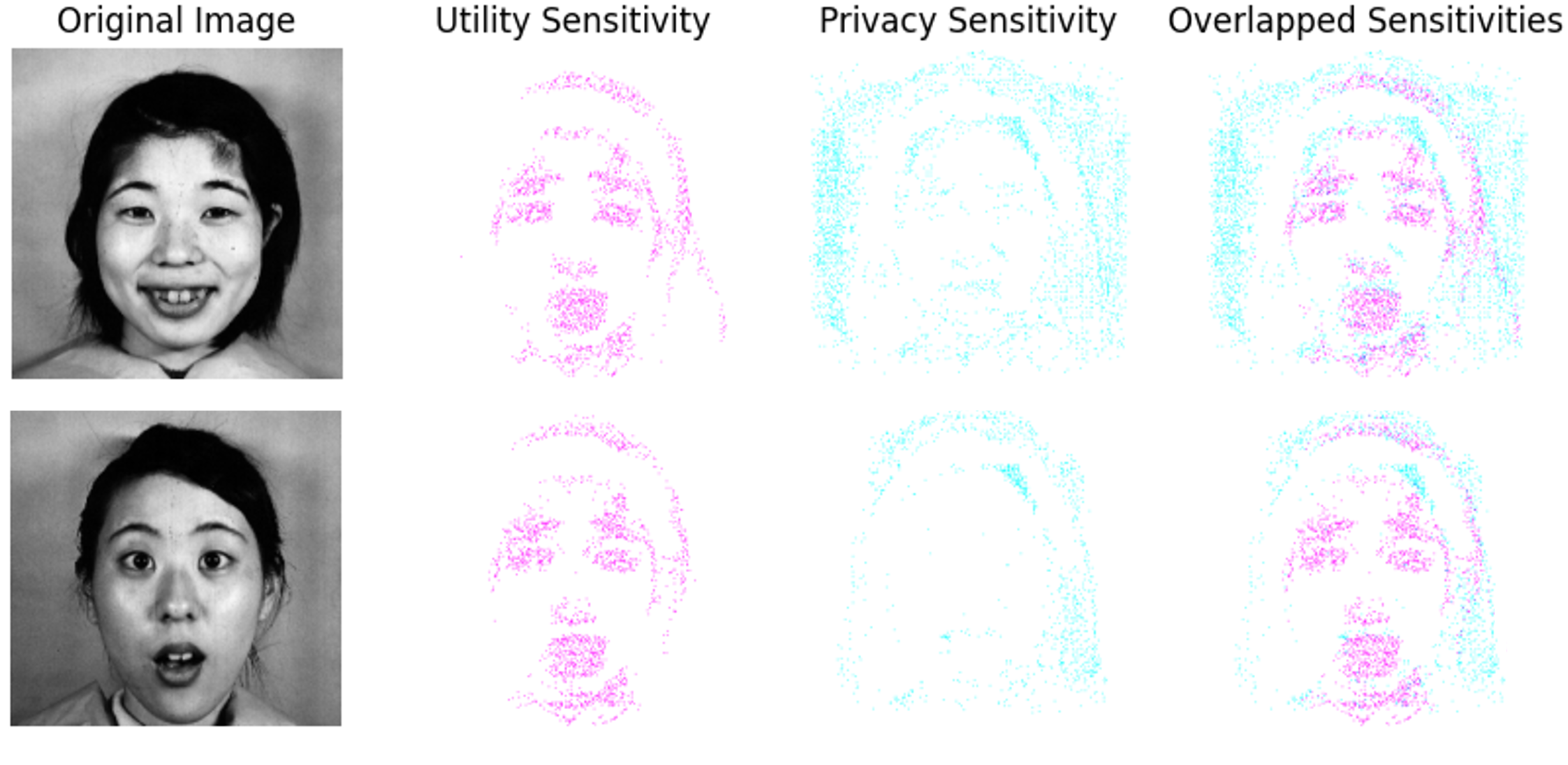} 
   \caption{An example of feature utility sensitivity and privacy sensitivity}
   \label{fig:sensitivities}
\end{figure}

However, as shown in Figure~\ref{fig:sensitivities}, it's likely to have features with both high utility and privacy sensitivity levels, e.g., some critical features in face images shared by both the expression classification and identity tasks. Thus achieving $u^T s = 0$ in real-world scenarios might be impossible. Instead, we can optimize the masking mechanism to maximize the utility with a desired amount of privacy preserved. Specifically, we formulate an optimization problem as follows to find a masking vector $m$:

\begin{equation*}
\begin{aligned}
& \underset{m}{\text{argmax}}
& & m^T u \\
& \text{s.t.} 
& & m^Ts < (1-\alpha) \sum_{j=1}^K s_j,
\end{aligned}
\end{equation*}

where $\alpha$ is the model owner's desired privacy preservation level and we use $\sum s_j$ to approximately represent the total amount of privacy information the features bring. The optimization will try to pick up the optimal mask maximizing the amount of utility, i.e., $m^Tu$, with acceptable privacy loss. This linear optimization problem is often solvable with the standard technique.

\begin{figure*}[htbp]
  \centering
  \includegraphics[width=0.9\linewidth, height=0.2\linewidth]{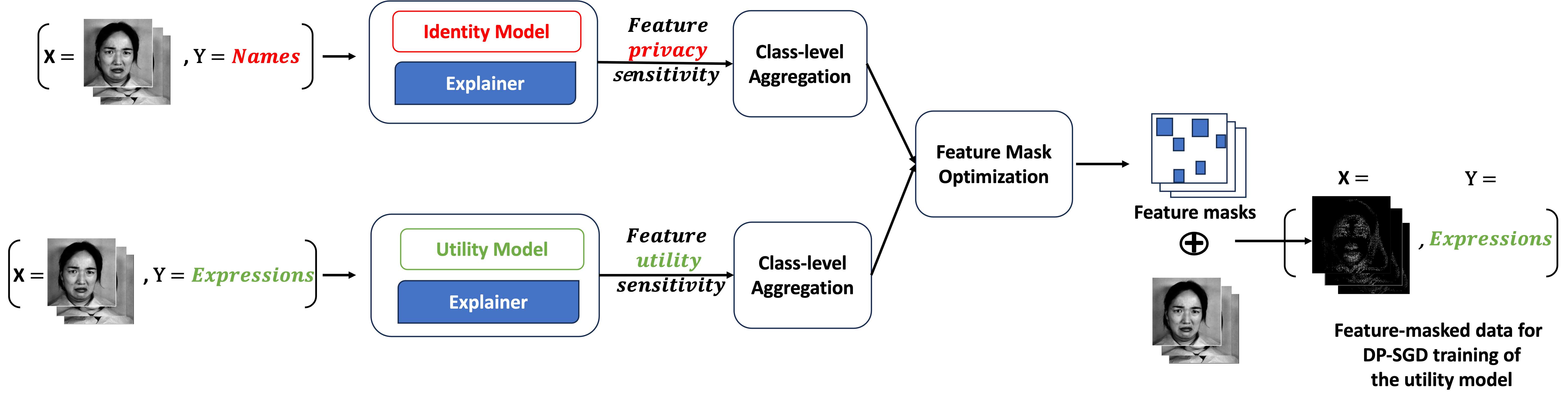}
  \caption{Pipeline of feature masking}
  \label{fig:pipeline}
\end{figure*}

\textbf{Implementation details.} As shown in Algorithm~\ref{alg:feature_masking} and Figure~\ref{fig:pipeline}, our implementation uses class-wise feature privacy sensitivity generation to preserve utility best and reduce privacy information during feature masking. Specifically, the data owner uses the model explainer to generate a privacy-sensitivity vector $s_{i,z}$ for a sample $x_{i,z}$ and utility-sensitivity vector $u_{i,y}$ for a sample $x_{i,y}$, where the same sample $x_i$ showing up in the two different tasks and labeled as class $z$ and $y$ from label set $Z$ and $Y$, respectively. By summing up the feature sensitivity vectors for all samples in one class, we get a class-wise feature privacy-sensitivity vector $S_z$ and utility-sensitivity vector $U_y$. In case the optimization problem is not solvable\footnote{This occurs with the probability of $3\%$ in our experiments.}, we use the top-k\% method instead. Specifically, we identify the features with top-k\% privacy-sensitivity and define the mask for each class. In the utility-oriented optimization method, we can also similarly derive a per-class mask. 

With a class-wise binary mask $m_z$ in place, we can process images from class $z$ to mask (or retain) their features accordingly: $x_{i,z}' = x_{i,z}^Tm_z$. 

\begin{algorithm}
\caption{Feature-masking mechanism}
\label{alg:feature_masking}
\begin{algorithmic}[1]
\Require Identity Model $I(D,Z)$, Image set $D$ of size $N_d$ with classes $1, 2, \dots, Z$, size $N_z$ of class $z$, Explainer $E()$
\Ensure Masked Image set $D'$
\Function{FeatureMasking}{$M(D,Y)$, $D$}
    \For{$i \gets 1, N_d$}
        \State $s_{i,z} \gets E(I(D,Z), x_{i,z})$ \Comment{Compute individual privacy-sensitivity vector}
        \State $u_{i,y} \gets E(M(D,Y), x_{i,y})$ \Comment{Compute individual utility-sensitivity vector}
    \EndFor
    \State $S_z \gets \frac{1}{N_z}\sum_{i=1}^{N_z} s_{i,z}$ 
     \State $U_y \gets \frac{1}{N_y}\sum_{i=1}^{N_y} s_{i,y}$ \Comment{Aggregate to class-wise}
    \State $m_z \gets 
        \begin{cases} 
            \text{Linear optimization} & \text{if solvable}\\
            \text{Top-k\% method} & \text{otherwise} 
        \end{cases}$ 
    \For{$i \gets 1, N_d$}
        \State $x_{i,z}' \gets x_{i,z}^Tm_z$ \Comment{Mask the images}
        \State Add $x_{i,z}'$ to $D'$
    \EndFor
    \State \Return $D'$
\EndFunction
\end{algorithmic}
\end{algorithm}

\section{Experiments}
\label{sec:exps}

We have conducted comprehensive experiments to answer the critical research questions in our study. 1) How does $ASR^M$ change over different datasets or versions of modified datasets? 2) With different versions of feature-masked data, how do the theoretical $\epsilon$ values differ at the same $ASR^M$ of utility models?  3) How do different feature masking strategies affect data utility on the utility models? 4) How does the choice of model explainer affect performance? And 5) What are the costs of the proposed methods?

\subsection{Setup}
\label{sec:expsetup}
\textbf{Datasets:} In addition to widely-used datasets such as MNIST, CIFAR10, and ImageNet-1K, we employ three facial expression datasets: RaFD \cite{langner2010presentation}, JAFFE \cite{lyons2021excavating}, and TFEID \cite{chen2007taiwanese}, along with a distracted driver activity recognition dataset, 100-Driver \cite{100Driver}, to demonstrate the performance of feature masking in practical scenarios. These datasets were selected because they clearly contain sensitive information and they intuitively show how two distinct sets of labels: utility labels and identity labels (i.e., the identities of the contributors) can be defined on the same dataset, corresponding to the utility and identity tasks. The utility labels represent 7 types of facial expressions in the facial expression datasets and 22 categories of distracted drivers' activities in the driver activity recognition dataset. Each dataset has also clearly defined which human subject each image belongs to, certainly without real identity information.  In training models, we split each dataset into training and testing sets using an 80:20 ratio.

Some preprocessing steps have been performed on these datasets. The RaFD dataset contains images of 67 subjects, each exhibiting 8 distinct emotional expressions. The JAFFE dataset includes 210 images of 10 female subjects, each displaying 7 expressions, repeated three times. The TFEID dataset consists of images from 40 subjects, each with 8 different facial expressions. Since the removal of sensitive features requires a model explainer to interpret each training sample, having too many classes can degrade explanation quality. To mitigate this, we randomly selected 10 identities from the RaFD and TFEID datasets, resulting in 80 images per dataset. The 100-Driver dataset includes 100 identities, each performing 22 distinct activity types. Each subject’s activity type contains 16 images of different settings, e.g., with/without wearing, multiple cameras with different angles in one vehicle, and different vehicles. In total, there are 470K images. Due to computational limitations, we randomly sample the same number of images for both identity and distraction labels, which generates a subset with 35K images. 

\textbf{Model training.} When using Differentially Private Stochastic Gradient Descent (DP-SGD) in training models, we adopted the following parameter settings: We use ResNet-101 for ImageNet-1K and 100-Driver\cite{100Driver} datasets for both the utility and identity models due to the higher complexity of these datasets, and ResNet-18 for other smaller datasets. All models are adjusted only in terms of input dimensions like image size and channel number to suit our dataset. A learning rate of 0.001 and epochs of 150 were set. We used a batch size of 5. Additionally, to prevent overfitting, an early stopping mechanism was implemented, halting training when no improvement was seen on the validation set for a set number of epochs.

\begin{table}[h]
\centering
\begin{tabular}{|c|c|c|c|c|} 
\hline
Dataset & Identity         & Utility           & \multicolumn{1}{c|}{Identities} & Size  \\ 
\hline
RaFD    & 0.737 (+/-0.031) & 0.293 (+/-0.027)  & \multirow{3}{*}{10} & 80       \\ 
\cline{1-3}\cline{5-5}
JAFFE   & 0.832 (+/-0.025) & 0.327 (+/-0.011)  &                     & 210      \\ 
\cline{1-3}\cline{5-5}
TFEID   & 0.719 (+/-0.028) & 0.286 (+/- 0.024) &                     & 80       \\
\cline{1-3}\cline{5-5}
100-Driver   & 0.913 (+/-0.012) & 0.721 (+/- 0.016) &                     & 35208       \\
\hline
\end{tabular}
\caption{Baseline models on the three datasets.}
\label{tab:baseline}
\end{table}

\textbf{Model Explainers.} To evaluate the impact of model explainers, we selected two of the most widely used methods: SHAP\footnote{\url{https://github.com/shap/shap}} and LIME\footnote{\url{https://github.com/marcotcr/lime}}. Both were employed to generate feature sensitivity scores for privacy and utility assessments. Since SHAP is more efficient for batch processing, we used it to address the first three questions. LIME, on the other hand, was applied to analyze the variations introduced by different model explainers.

\textbf{Evaluation metrics.} The $ASR^M$ values are generated with the hypothesis testing method described in Section~\ref{sec:DP_Auditing}. $ASR^M$ is the maximum sample $ASR $ over all samples in a dataset to be consistent with the worst-case definition of differential privacy. $ASR^{M,\epsilon}$ is used to indicate $ASR^M$ on DP-SGD trained models. We use model accuracy to evaluate the utility models.

\textbf{Deriving $ASR^M$.} To conduct the LiRA, we train 1000 shadow models\footnote{Training 1000 ResNet-18 models on an NVIDIA TITAN V100 averagely spend 19 hours, and ResNet-101 spend 42 hours} -- for each shadow model we flip coins for each sample to split the dataset into in-domain and out-domain samples, which are used to train the shadow model.  Then, we apply the offline version of LiRA \cite{carlini2022membership} to estimate the $ASR^M$. The smaller the $ASR^M$, the more difficult it is to distinguish samples, and thus the privacy is better preserved.

\subsection{Result Analysis: $ASR^M$ for Modified Datasets}

\begin{figure}[htbp]
  \centering
  \begin{subfigure}{0.24\textwidth}
    \centering
    \begin{tikzpicture}[scale=0.8]
      \begin{axis}[
        xlabel={Feature Masked (\%)},
        ylabel={$ASR^{M_{\epsilon=8}}$},
        ymin = 0.48,
        ymax=0.7,
        xmin = -1,
        xmax = 21,
        ytick={0.5, 0.55, 0.6, 0.65, 0.7},
        ymajorgrids=true,
            grid style=dashed,
        legend entries={C-10, IN},
        scale only axis,
        width=0.95\textwidth,
        height=0.6\textwidth,
        enlarge x limits = false,
        enlarge y limits = false,
        legend style={
          at={(0.6, 1.0)},
          anchor=north,
          legend columns=-11,
          /tikz/every even column/.append style={column sep=0.5cm}
        },
        x tick label style={
        rotate=0,
        inner sep=0pt, 
        xshift=0pt 
    } ,
     y tick label style={
        rotate=0,
        inner sep=0pt, 
        xshift=-2pt
    } 
      ]
      \addplot[color=blue, mark=o] coordinates {
        (0,0.615)(5,0.593)(10,0.564)(15,0.542)(20,0.526)
      };
      \addplot[color=red, mark=o] coordinates {
        (0,0.613)(5,0.603)(10,0.573)(15,0.553)(20,0.534)
      };
      \end{axis}
    \end{tikzpicture}
    \caption{$ASR^M$ drops with more removed features.(C-10,IN). $\epsilon=8$}
    \label{f1d}
  \end{subfigure}
  \hfill
   \begin{subfigure}{0.24\textwidth}
    \centering
    \begin{tikzpicture}[scale=0.8]
      \begin{axis}[
          xlabel={Labels Randomized (\%)},
          width=0.95\textwidth,
          ymajorgrids=true,
          grid style=dashed,
          height=0.6\textwidth,
          xmin=-4, xmax=104,
          ymin=0.48, ymax=0.7,
          xtick={0,20,40,60,80,100},        
          scale only axis,
          ymajorgrids=true,
          grid style=dashed,
           x tick label style={
        rotate=0,
        inner sep=0pt, 
        xshift=0pt 
    } ,
     y tick label style={
        rotate=0,
        inner sep=0pt, 
        xshift=-2pt 
    } 
        ]
        \addplot[color=black, mark=o,] coordinates {
          (0,0.622) 
          (10,0.613) 
          (20,0.574) 
          (30,0.563) 
          (40,0.543) 
          (50,0.532) 
          (60,0.512) 
          (70,0.515) 
          (80,0.503) 
          (90,0.506) 
          (100,0.504)
        };
      \end{axis}
    \end{tikzpicture}
    \caption{$ASR^M$ drops with more labels are randomized(MNIST).}
    \label{f1b}
  \end{subfigure}
  \caption{Figure (a) shows the decreasing trends of $ASR^M$ when top-k important features are masked for CIFAR-10 and ImageNet-1K. Figure (b) demonstrates that with the increasing percentage of randomized labels (representing data quality reduction), $ASR^M$ decreases for MNIST. }
  \label{fig:performance_epsilon}
\end{figure}
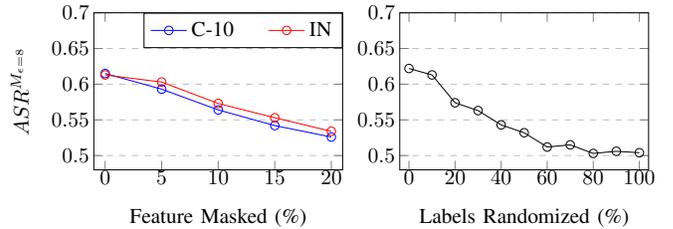

\begin{figure*}[htbp]
  \centering
   \begin{subfigure}[t]{0.24\textwidth}
     \centering
    \begin{tikzpicture}[scale = 0.7]
      \begin{axis}[
        ylabel={$\epsilon$},
        xtick={0.5, 0.55, 0.6, 0.65, 0.7},
         xmin=0.5, xmax=0.7,    
        xlabel={$ASR^{M_\epsilon}$},
        grid=both,
        ylabel near ticks,
        scale only axis,
        width=\linewidth
      ]
	\addplot[color=red, mark=o,smooth] coordinates {
 	 (0.517, 0.5) (0.525, 1) (0.531, 1.5) (0.556, 2.0) (0.572, 2.5) (0.584, 3.0) (0.617, 3.5) (0.625, 4.0) (0.631, 4.5) (0.636, 5.0) (0.634, 5.5) (0.643, 6.0) (0.652, 6.5) (0.657, 7.0)
	};
	\addplot[color=black, mark=o, smooth] coordinates {
 	 (0.512, 0.5) (0.513, 1) (0.518, 1.5) (0.521, 2.0) (0.526, 2.5) (0.533, 3.0) (0.545, 3.5) (0.553, 4.0) (0.567, 4.5) (0.571, 5.0) (0.589, 5.5) (0.612, 6.0) (0.613, 6.5) (0.619, 7.0)
	};
	\addplot[color=blue, mark=o,smooth] coordinates {
 	 (0.504, 0.5) (0.512, 1) (0.514, 1.5) (0.517, 2.0) (0.521, 2.5) (0.524, 3.0) (0.522, 3.5) (0.526, 4.0) (0.531, 4.5) (0.537, 5.0) (0.534, 5.5) (0.535, 6.0) (0.533, 6.5) (0.542, 7.0)
	};
	\addplot[color=black, dotted, thick] coordinates {
	(0.54, 0) (0.54,8)
	};
      \end{axis}
    \end{tikzpicture}
    \caption{JAFFE (SHAP)}
    \label{JAFFE_R}
  \end{subfigure}
   \begin{subfigure}[t]{0.24\textwidth}
     \centering
    \begin{tikzpicture}[scale = 0.7]
      \begin{axis}[
        ylabel={$\epsilon$},
        xtick={0.5, 0.55, 0.6, 0.65, 0.7},
         xmin=0.5, xmax=0.7,    
        xlabel={$ASR^{M_\epsilon}$},
        grid=both,
        scale only axis,
        width=\linewidth
      ]
        \addplot[color=red, mark=o,smooth] coordinates {
 	 (0.526, 0.5) (0.531, 1) (0.536, 1.5) (0.551, 2.0) (0.568, 2.5) (0.571, 3.0) (0.589, 3.5) (0.612, 4.0) (0.619, 4.5) (0.621, 5.0) (0.632, 5.5) (0.641, 6.0) (0.639, 6.5) (0.638, 7.0)
	};
	\addplot[color=black, mark=o, smooth] coordinates {
 	 (0.518, 0.5) (0.526, 1.5) (0.530, 2.0) (0.535, 2.5) (0.539, 3.0) (0.544, 3.5) (0.551, 4.0) (0.553, 4.5) (0.567, 5.0) (0.594, 5.5) (0.621, 6.0) (0.619, 6.5) (0.624, 7.0)
	};
	\addplot[color=blue, mark=o,smooth] coordinates {
 	 (0.515, 0.5) (0.521, 1.5) (0.527, 2.0) (0.533, 2.5) (0.536, 3.0) (0.535, 3.5) (0.537, 4.0) (0.54, 4.5) (0.542, 5.0) (0.539, 5.5) (0.542, 6.0) (0.541, 6.5) (0.539, 7.0)
	};
	\addplot[color=black, dotted, thick] coordinates {
	(0.54, 0) (0.54,8)
	};
      \end{axis}
    \end{tikzpicture}
    \caption{RaFD (SHAP)}
  \end{subfigure}
   \begin{subfigure}[t]{0.24\textwidth}
     \centering
    \begin{tikzpicture}[scale = 0.7]
      \begin{axis}[
        ylabel={$\epsilon$},
        xtick={0.5, 0.55, 0.6, 0.65, 0.7},
         xmin=0.5, xmax=0.7,    
        xlabel={$ASR^{M_\epsilon}$},
        grid=both,
        legend pos=north west,
        scale only axis,
        width=\linewidth
      ]
	\addplot[color=red, mark=o, smooth] coordinates {
  	(0.525, 0.5) (0.527, 1) (0.539, 1.5) (0.551, 2.0) (0.567, 2.5) (0.571, 3.0) (0.588, 3.5) (0.592, 4.0) (0.612, 4.5) (0.614, 5.0) (0.624, 5.5) (0.627, 6.0) (0.625, 6.5) (0.628, 7.0)
	};
	\addplot[color=black, mark=o, smooth] coordinates {
	  (0.522, 0.5) (0.524, 1) (0.528, 1.5) (0.533, 2.0) (0.539, 2.5) (0.542, 3.0) (0.547, 3.5) (0.551, 4.0) (0.572, 4.5) (0.577, 5.0) (0.585, 5.5) (0.592, 6.0) (0.603, 6.5) (0.609, 7.0)
	};
	\addplot[color=blue, mark=o, smooth] coordinates {
	  (0.516, 0.5) (0.519, 1) (0.523, 1.5) (0.526, 2.0) (0.525, 2.5) (0.531, 3.0) (0.534, 3.5) (0.537, 4.0) (0.536, 4.5) (0.542, 5.0) (0.546, 5.5) (0.547, 6.0) (0.552, 6.5) (0.551, 7.0)
	};
	\addplot[color=black, dotted, thick] coordinates {
  	(0.54, 0) (0.54,8)
	};
      \end{axis}
    \end{tikzpicture}
    \caption{TFEID (SHAP)}
  \end{subfigure}
   \begin{subfigure}[t]{0.24\textwidth}
     \centering
    \begin{tikzpicture}[scale = 0.7]
      \begin{axis}[
        ylabel={$\epsilon$},
        xtick={0.5, 0.55, 0.6, 0.65, 0.7},
         xmin=0.5, xmax=0.7,    
        xlabel={$ASR^{M_\epsilon}$},
        grid=both,
        scale only axis,
         legend entries={Original, Random FM, Optimized FM},
         legend style={
          at={(0.64, 0.37)},
          anchor=north,
          legend columns=1,
          /tikz/every even column/.append style={column sep=0.5cm}
        },
        width=\linewidth
      ]
        \addplot[color=red, mark=o,smooth] coordinates {
 	 (0.513, 0.5) (0.512, 1) (0.529, 1.5) (0.534, 2.0) (0.553, 2.5) (0.558, 3.0) (0.562, 3.5) (0.588, 4.0) (0.614, 4.5) (0.616, 5.0) (0.621, 5.5) (0.626, 6.0) (0.619, 6.5) (0.621, 7.0)
	};
	\addplot[color=black, mark=o, smooth] coordinates {
 	 (0.505, 0.5) (0.511, 1.5) (0.514, 2.0) (0.522, 2.5) (0.526, 3.0) (0.531, 3.5) (0.546, 4.0) (0.543, 4.5) (0.547, 5.0) (0.554, 5.5) (0.563, 6.0) (0.571, 6.5) (0.573, 7.0)
	};
	\addplot[color=blue, mark=o,smooth] coordinates {
 	 (0.511, 0.5) (0.516, 1.5) (0.521, 2.0) (0.527, 2.5) (0.524, 3.0) (0.526, 3.5) (0.531, 4.0) (0.528, 4.5) (0.533, 5.0) (0.531, 5.5) (0.527, 6.0) (0.532, 6.5) (0.536, 7.0)
	};
	\addplot[color=black, dotted, thick] coordinates {
	(0.54, 0) (0.54,8)
	};
      \end{axis}
    \end{tikzpicture}
    \caption{100-Driver(SHAP)}
    \label{fig:100Driver(SHAP)}
  \end{subfigure}
\caption{The relationship between $ASR^{M, \epsilon}$ and theoretical $\epsilon$ over utility models for optimized masked, random masked, and original dataset. The $\alpha$ parameter setting for optimized feature masking: $\alpha$ at 0.1 for JAFFE, 0.2 for both RaFD and TFEID, and 0.3 for 100-Driver as suggested later in Figure~\ref{fig:choice of alpha}. For the random feature masking approach, we omit 30\% of the features at random, i.e., the same number of masked features in the optimized feature masking setting.}
  \label{fig: Differential Privacy: privacy}
\end{figure*}
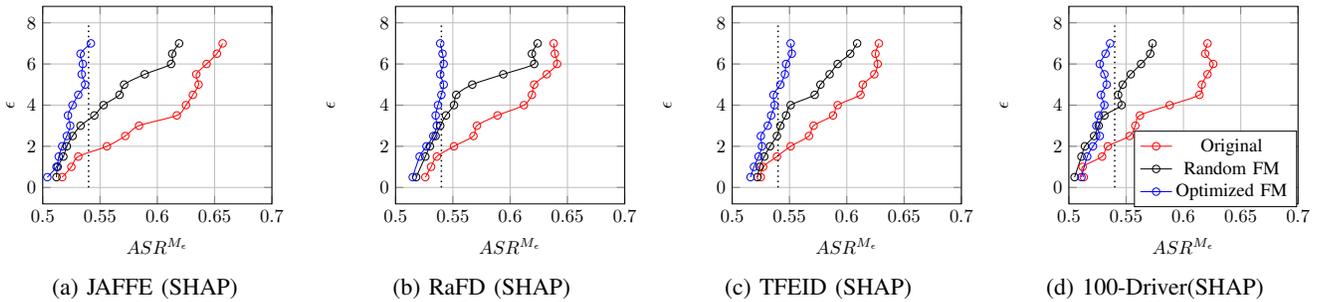

The first goal of our approach is to identify a data- and model-specific measure that can evaluate the practical privacy risk. We have shown some examples earlier in Figure \ref{f1c} that $ASR^M$ can serve this purpose well. Here, we show more detailed experiments to see how different dataset qualities may affect $ASR^M$. In Figure~\ref{fig:performance_epsilon}, we take the well-known datasets CIFAR-10 and ImageNet-1K and progressively mask the top-ranked features (pixels), and their $ASR^M$ drops correspondingly. To understand how label quality may affect $ASR^M$, we also randomize a portion of MNIST's label, and we have observed a similar decreasing trend. These results motivated us to explore the feature masking approach to change the dataset sensitivity to achieve better utility and privacy balances.

\subsection{Result Analysis: $ASR^{M, \epsilon}$ vs theoretical $\epsilon$}
\label{sec:results_Analysis_2}
To understand the correlation between $\epsilon$ and $ASR^{M, \epsilon}$, we take the three versions of datasets for experiments: the original (Original), the dataset with 30\% randomly masked features (Random FM), and that with optimized feature masking (Optimal FM). Each point in Figures \ref{fig: Differential Privacy: privacy} represents a set of experiments: we apply DP-SGD with the specific theoretical $\epsilon$ setting to train a set of identity models and then use the LiRA to derive the corresponding $ASR^M$. 
The result in Figure~\ref{fig: Differential Privacy: privacy} shows that (1) $ASR^M$ is positively correlated with the theoretical $\epsilon$. However, the correlation is stronger for the original data. Due to the reduced privacy sensitivity, the FM methods have a narrower range of $ASR^M$. (2) The optimized FM has advantages in significantly lower $ASR^M$, which implies reduced practical privacy risk. (3) For the same level of $ASR^M$, we can probably use a much higher theoretical $\epsilon$ for DP-SGD. For example, for JAFFE, $\epsilon=1$ gives $ASR^M$ around 0.525 (the 2nd red dot from left to right on the DP-SGD curve in Figure~\ref{JAFFE_R}. With the same $ASR^M$, the $\epsilon$ for the optimized feature-masked dataset can be relaxed to around 4. The relaxed $\epsilon$ will allow for preserving more useful information for the utility models, which will be examined next. 

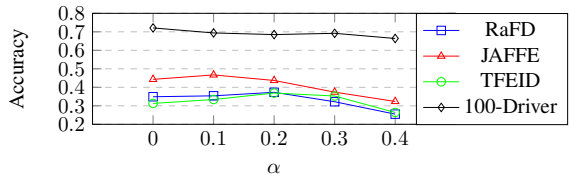
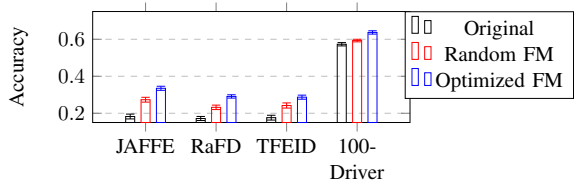
\begin{figure}[htbp]
    \centering
    \captionsetup{justification=centering}
    \begin{subfigure}[t]{0.42\textwidth}
        \centering
        \begin{tikzpicture}[scale=0.8]
        \begin{axis}[
            width=\textwidth,  
            height=0.45\textwidth,
            xlabel={$\alpha$},
            ylabel={Accuracy},
            xmin=-0.1, xmax=0.5,
            ymin=0.2, ymax=0.8,
            xtick={0,0.1,0.2,0.3,0.4},
            ytick={0.1,0.2,0.3,0.4,0.5,0.6,0.7,0.8},
            legend style={at={(1.1,1.002)},anchor=north,legend columns=1},
            ymajorgrids=true,
            grid style=dashed
        ]
        \addplot[color=blue, mark=square,] coordinates {(0,0.349) (0.1,0.354) (0.2,0.373) (0.3,0.322) (0.4,0.255)};
        \addlegendentry{RaFD}
        \addplot[color=red, mark=triangle,] coordinates {(0,0.443) (0.1,0.467) (0.2,0.437) (0.3,0.374) (0.4,0.3235)};
        \addlegendentry{JAFFE}
        \addplot[color=green, mark=o,] coordinates {(0,0.313) (0.1,0.334) (0.2,0.369) (0.3,0.353) (0.4,0.264)};
        \addlegendentry{TFEID}
        \addplot[color=black, mark=diamond,] coordinates {(0,0.721) (0.1,0.694) (0.2,0.685) (0.3,0.692) (0.4,0.664)};
        \addlegendentry{100-Driver}
        \end{axis}
        \end{tikzpicture}
        \caption{Accuracy of utility models without DP-SGD across different $\alpha$ values.}
        \label{fig:choice of alpha}
    \end{subfigure}
    \hfill
        \begin{subfigure}[t]{0.42\textwidth}
        \centering
        \begin{tikzpicture}[scale=0.8]
        \begin{axis}[
            ybar=3pt,
   	    bar width=.15cm,
            width=0.9\textwidth,  
            height=0.45\textwidth,
            legend style={at={(1.25,1.002)},anchor=north,legend columns=1},
            ylabel={Accuracy},
            symbolic x coords={JAFFE, RaFD, TFEID,100-Driver},
            xtick=data,
            enlarge x limits=0.25,
            ymin=0.15, ymax=0.75,
            ymajorgrids=true,
            grid style=dashed,
            x tick label style={text width=1.5cm,align=center}
        ]
        \addplot[black, error bars/.cd, y dir=both, y explicit] 
        coordinates {
            (JAFFE,0.1824) +- (0,0.013)
            (RaFD,0.1717) +- (0,0.011)
            (TFEID,0.1762) +- (0,0.013)
            (100-Driver,0.5731) +- (0,0.009)
        };
        \addplot[red, error bars/.cd, y dir=both, y explicit] 
        coordinates {
            (JAFFE,0.2733) +- (0,0.013)
            (RaFD,0.232) +- (0,0.012)
            (TFEID,0.242) +- (0,0.014)
            (100-Driver,0.593) +- (0,0.007)
        };
        \addplot[blue, error bars/.cd, y dir=both, y explicit] 
        coordinates {
            (JAFFE,0.335) +- (0,0.011) 
            (RaFD,0.291) +- (0,0.009)
            (TFEID,0.287) +- (0,0.011)
            (100-Driver,0.637) +- (0,0.009)
        };
        \legend{Original, Random FM, Optimized FM}
        \end{axis}
        \end{tikzpicture}
        \caption{Performance of the utility model with the DP-SGD and $\epsilon$ values at $ASR^M = 0.54$, detailed in Table~\ref{tab:summary}}
         \label{fig:Differential Privacy: acc}
    \end{subfigure}%
    \caption{FM optimization algorithm helps preserve better data utility}
    \label{fig:combined}
\end{figure}

 \begin{figure*}[htbp]
  \centering
   \begin{subfigure}[t]{0.24\textwidth}
     \centering
    \begin{tikzpicture}[scale = 0.7]
      \begin{axis}[
        ylabel={$\epsilon$},
        xtick={0.5, 0.55, 0.6, 0.65, 0.7},
         xmin=0.5, xmax=0.7,    
        xlabel={$ASR^{M_\epsilon}$},
        grid=both,
        ylabel near ticks,
        scale only axis,
        width=\linewidth
      ]
	\addplot[color=red, mark=o,smooth] coordinates {
 	 (0.517, 0.5) (0.531, 1.5) (0.556, 2.0) (0.572, 2.5) (0.584, 3.0) (0.617, 3.5) (0.625, 4.0) (0.631, 4.5) (0.636, 5.0) (0.634, 5.5) (0.643, 6.0) (0.652, 6.5) (0.657, 7.0)
	};
	\addplot[color=black, mark=o, smooth] coordinates {
 	 (0.517, 0.5) (0.522, 1.5) (0.525, 2.0) (0.524, 2.5) (0.531, 3.0) (0.537, 3.5) (0.549, 4.0) (0.562, 4.5) (0.574, 5.0) (0.585, 5.5) (0.596, 6.0) (0.609, 6.5) (0.614, 7.0)
	};
	\addplot[color=blue, mark=o,smooth] coordinates {
 	 (0.511, 0.5) (0.521, 1.5) (0.524, 2.0) (0.527, 2.5) (0.528, 3.0) (0.531, 3.5) (0.533, 4.0) (0.535, 4.5) (0.538, 5.0) (0.533, 5.5) (0.537, 6.0) (0.534, 6.5) (0.542, 7.0)
	};
	\addplot[color=black, dotted, thick] coordinates {
	(0.54, 0) (0.54,8)
	};
      \end{axis}
    \end{tikzpicture}
    \caption{JAFFE (LIME)}
    \label{JAFFE_R}
  \end{subfigure}
   \begin{subfigure}[t]{0.24\textwidth}
     \centering
    \begin{tikzpicture}[scale = 0.7]
      \begin{axis}[
        ylabel={$\epsilon$},
        xtick={0.5, 0.55, 0.6, 0.65, 0.7},
         xmin=0.5, xmax=0.7,    
        xlabel={$ASR^{M_\epsilon}$},
        grid=both,
        scale only axis,
        width=\linewidth
      ]
        \addplot[color=red, mark=o,smooth] coordinates {
 	 (0.526, 0.5) (0.531, 1) (0.536, 1.5) (0.551, 2.0) (0.568, 2.5) (0.571, 3.0) (0.589, 3.5) (0.612, 4.0) (0.619, 4.5) (0.621, 5.0) (0.632, 5.5) (0.641, 6.0) (0.639, 6.5) (0.638, 7.0)
	};
	\addplot[color=black, mark=o, smooth] coordinates {
 	 (0.520, 0.5) (0.524, 1.5) (0.529, 2.0) (0.527, 2.5) (0.534, 3.0) (0.539, 3.5) (0.546, 4.0) (0.551, 4.5) (0.561, 5.0) (0.572, 5.5) (0.594, 6.0) (0.606, 6.5) (0.616, 7.0)
	};
	\addplot[color=blue, mark=o,smooth] coordinates {
 	 (0.519, 0.5) (0.524, 1.5) (0.531, 2.0) (0.535, 2.5) (0.532, 3.0) (0.537, 3.5) (0.541, 4.0) (0.546, 4.5) (0.544, 5.0) (0.547, 5.5) (0.553, 6.0) (0.556, 6.5) (0.561, 7.0)
	};
	\addplot[color=black, dotted, thick] coordinates {
	(0.54, 0) (0.54,8)
	};
      \end{axis}
    \end{tikzpicture}
    \caption{RaFD (LIME)}
  \end{subfigure}
   \begin{subfigure}[t]{0.24\textwidth}
     \centering
    \begin{tikzpicture}[scale = 0.7]
      \begin{axis}[
        ylabel={$\epsilon$},
        xtick={0.5, 0.55, 0.6, 0.65, 0.7},
         xmin=0.5, xmax=0.7,    
        xlabel={$ASR^{M_\epsilon}$},
        grid=both,
        legend pos=north west,
        scale only axis,
        width=\linewidth
      ]
	\addplot[color=red, mark=o, smooth] coordinates {
  	(0.525, 0.5) (0.527, 1) (0.539, 1.5) (0.551, 2.0) (0.567, 2.5) (0.571, 3.0) (0.588, 3.5) (0.592, 4.0) (0.612, 4.5) (0.614, 5.0) (0.624, 5.5) (0.627, 6.0) (0.625, 6.5) (0.628, 7.0)
	};
	\addplot[color=black, mark=o, smooth] coordinates {
	  (0.526, 0.5) (0.532, 1) (0.534, 1.5) (0.538, 2.0) (0.533, 2.5) (0.539, 3.0) (0.542, 3.5) (0.548, 4.0) (0.559, 4.5) (0.564, 5.0) (0.579, 5.5) (0.59, 6.0) (0.607, 6.5) (0.612, 7.0)
	};
	\addplot[color=blue, mark=o, smooth] coordinates {
	  (0.521, 0.5) (0.526, 1) (0.531, 1.5) (0.53, 2.0) (0.533, 2.5) (0.537, 3.0) (0.539, 3.5) (0.533, 4.0) (0.543, 4.5) (0.547, 5.0) (0.551, 5.5) (0.546, 6.0) (0.549, 6.5) (0.555, 7.0)
	};
	\addplot[color=black, dotted, thick] coordinates {
  	(0.54, 0) (0.54,8)
	};
      \end{axis}
    \end{tikzpicture}
    \caption{TFEID (LIME)}
  \end{subfigure}
     \begin{subfigure}[t]{0.24\textwidth}
     \centering
    \begin{tikzpicture}[scale = 0.7]
      \begin{axis}[
        ylabel={$\epsilon$},
        xtick={0.5, 0.55, 0.6, 0.65, 0.7},
         xmin=0.5, xmax=0.7,    
        xlabel={$ASR^{M_\epsilon}$},
        grid=both,
        scale only axis,
         legend entries={Original, Random FM, Optimized FM},
         legend style={
          at={(0.64, 0.37)},
          anchor=north,
          legend columns=1,
          /tikz/every even column/.append style={column sep=0.5cm}
        },
        width=\linewidth
      ]
        \addplot[color=red, mark=o,smooth] coordinates {
 	 (0.513, 0.5) (0.512, 1) (0.529, 1.5) (0.534, 2.0) (0.553, 2.5) (0.558, 3.0) (0.562, 3.5) (0.588, 4.0) (0.614, 4.5) (0.616, 5.0) (0.621, 5.5) (0.626, 6.0) (0.619, 6.5) (0.621, 7.0)
	};
	\addplot[color=black, mark=o, smooth] coordinates {
 	 (0.505, 0.5) (0.511, 1.5) (0.514, 2.0) (0.522, 2.5) (0.526, 3.0) (0.531, 3.5) (0.546, 4.0) (0.543, 4.5) (0.547, 5.0) (0.554, 5.5) (0.563, 6.0) (0.571, 6.5) (0.573, 7.0)
	};
	\addplot[color=blue, mark=o,smooth] coordinates {
 	 (0.511, 0.5) (0.516, 1.5) (0.523, 2.0) (0.527, 2.5) (0.525, 3.0) (0.536, 3.5) (0.538, 4.0) (0.542, 4.5) (0.543, 5.0) (0.549, 5.5) (0.556, 6.0) (0.567, 6.5) (0.569, 7.0)
	};
	\addplot[color=black, dotted, thick] coordinates {
	(0.54, 0) (0.54,8)
	};
      \end{axis}
    \end{tikzpicture}
    \caption{100-Driver(LIME)}
    \label{fig:100Driver(LIME)}
    \end{subfigure}
\caption{We use LIME to reproduce the results. The relationship between $ASR^{M_\epsilon}$ and theoretical $\epsilon$ over utility models for optimized masked, random masked, and original datasets show a similar trend with using SHAP (Figure~\ref{fig: Differential Privacy: privacy}). Parameter setting for optimized feature masking: $\alpha$ at 0.1 for both 100-Driver and JAFFE, 0.2 for both RaFD and TFEID, as suggested later in Figure~\ref{fig:choice of alpha_LIME}. For the random feature masking approach, we omit 30\% of the features at random, which approximates the number of masked features by optimized feature masking.}
  \label{fig: Differential Privacy: privacy_LIME}
\end{figure*}
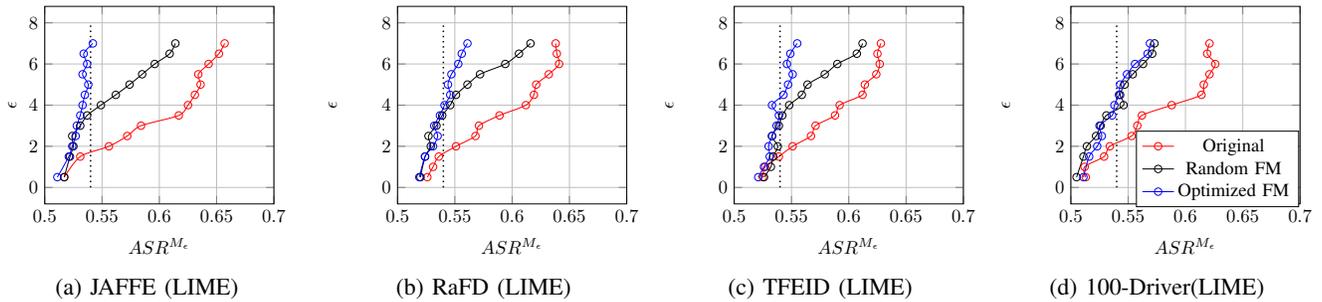

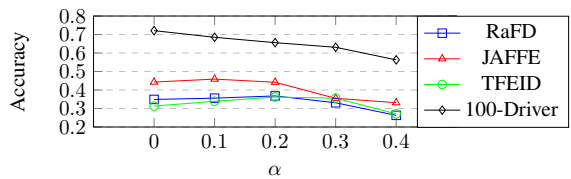
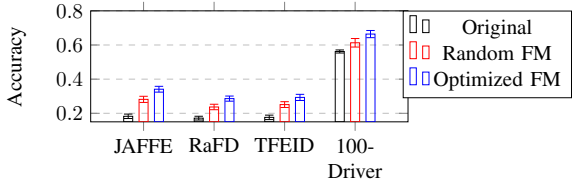
\begin{figure}[htbp]
    \centering
    \captionsetup{justification=centering}
    \begin{subfigure}[t]{0.42\textwidth}
        \centering
        \begin{tikzpicture}[scale=0.8]
        \begin{axis}[
            width=\textwidth,  
            height=0.45\textwidth,
            xlabel={$\alpha$},
            ylabel={Accuracy},
            xmin=-0.1, xmax=0.5,
            ymin=0.2, ymax=0.8,
            xtick={0,0.1,0.2,0.3,0.4},
            ytick={0.1,0.2,0.3,0.4,0.5,0.6,0.7,0.8},
            legend style={at={(1.1,1.002)},anchor=north,legend columns=1},
            ymajorgrids=true,
            grid style=dashed
        ]
        \addplot[color=blue, mark=square,] coordinates {(0,0.349) (0.1,0.356) (0.2,0.368) (0.3,0.331) (0.4,0.263)};
        \addlegendentry{RaFD}
        \addplot[color=red, mark=triangle,] coordinates {(0,0.443) (0.1,0.459) (0.2,0.442) (0.3,0.355) (0.4,0.3315)};
        \addlegendentry{JAFFE}
        \addplot[color=green, mark=o,] coordinates {(0,0.313) (0.1,0.339) (0.2,0.362) (0.3,0.356) (0.4,0.271)};
        \addlegendentry{TFEID}
        \addplot[color=black, mark=diamond,] coordinates {(0,0.721) (0.1,0.685) (0.2,0.656) (0.3,0.631) (0.4,0.563)};
        \addlegendentry{100-Driver}
        \end{axis}
        \end{tikzpicture}
        \caption{Accuracy of utility models without DP-SGD across different $\alpha$ values.}
        \label{fig:choice of alpha_LIME}
    \end{subfigure}
        \begin{subfigure}[t]{0.42\textwidth}
        \centering
        \begin{tikzpicture}[scale=0.8]
        \begin{axis}[
            ybar=3pt,
            bar width=.15cm,
            width=0.9\textwidth,  
            height=0.45\textwidth,
            legend style={at={(1.25,1.002)},anchor=north,legend columns=1},
            ylabel={Accuracy},
            symbolic x coords={JAFFE, RaFD, TFEID,100-Driver},
            xtick=data,
            ymajorgrids=true,
            grid style=dashed,
            enlarge x limits=0.25,
            ymin=0.15, ymax=0.8,
            x tick label style={text width=1.5cm,align=center}
        ]
        \addplot[black, error bars/.cd, y dir=both, y explicit] 
        coordinates {
            (JAFFE,0.1824) +- (0,0.013)
            (RaFD,0.1717) +- (0,0.011)
            (TFEID,0.1762) +- (0,0.013)
            (100-Driver,0.562) +- (0,0.009)
        };
        \addplot[red, error bars/.cd, y dir=both, y explicit] 
        coordinates {
            (JAFFE,0.2816) +- (0,0.018)
            (RaFD,0.237) +- (0,0.016)
            (TFEID,0.251) +- (0,0.017)
            (100-Driver,0.613) +- (0,0.025)
        };
        
        \addplot[blue, error bars/.cd, y dir=both, y explicit] 
        coordinates {
            (JAFFE,0.341) +- (0,0.017) 
            (RaFD,0.286) +- (0,0.015)
            (TFEID,0.293) +- (0,0.018)
            (100-Driver,0.664) +- (0,0.021)
        };
        \legend{Original, Random FM, Optimized FM}
        \end{axis}
        \end{tikzpicture}
        \caption{Performance of the utility model at $ASR \approx 0.54$, depicted as a black dotted line in Figure~\ref{fig: Differential Privacy: privacy_LIME}.}
        \label{fig:Differential Privacy: acc_Lime}
    \end{subfigure}%
    \caption{LIME-based FM optimization algorithm also helps preserve better data utility}
    \label{fig:combined_LIME}
\end{figure}

\begin{figure}
\centering
\begin{tikzpicture}[scale=0.8]
    \begin{axis}[
        ybar,
        bar width=.15cm,
        width=0.46\textwidth,
        height=0.22\textwidth,
        symbolic x coords={JAFFE, RaFD, TFEID, 100-Driver},
        xtick=data,
        ylabel={$ASR^{M_{\epsilon=7}}$},
        ymin=0.52, ymax=0.63,
        ymajorgrids=true,
        grid style=dashed,
        enlarge x limits=0.2,
        legend style={at={(1.1,1.0)}, anchor=north, legend columns=1},
        error bars/.cd
    ]
    
    \addplot+[error bars/.cd, y dir=both, y explicit] coordinates {
        (JAFFE, 0.539) +- (0, 0.004)
        (RaFD, 0.554) +- (0, 0.007)
        (TFEID, 0.548) +- (0, 0.005)
        (100-Driver, 0.536) +- (0, 0.008)
    };
    
    \addplot+[error bars/.cd, y dir=both, y explicit] coordinates {
        (JAFFE, 0.542) +- (0, 0.015)
        (RaFD, 0.561) +- (0, 0.013)
        (TFEID, 0.555) +- (0, 0.018)
        (100-Driver, 0.569) +- (0, 0.023)
    };
    \legend{SHAP, LIME}
    \end{axis}
\end{tikzpicture}
    \caption{$ASR^M$ when training utility model with $\epsilon=7$ and applying optimized FM. SHAP is more stable than LIME.}
    \label{fig:stability}
\end{figure}
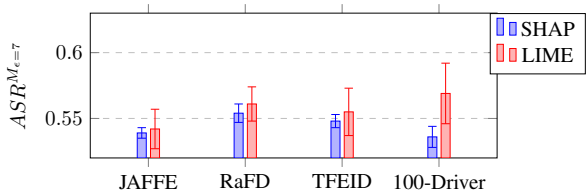

\subsection{Result Analysis: Optimizing Feature Masking}

As discussed in Section \ref{sec:optimization}, selecting an appropriate $\alpha$ for generating feature masks is critical. Here, $\alpha$ represents the extent of dataset utility sensitivity to be reduced. Figure~\ref{fig:choice of alpha} illustrates how the performance of the utility model changes with different $\alpha$ settings. Interestingly, increasing $\alpha$ initially enhances the utility model’s performance, but beyond a certain threshold, it begins to decline for the facial expression datasets. Specifically, we observe that the utility model achieves the best accuracy at $\alpha = 0.2$ for RaFD and TFEID, and at $\alpha = 0.1$ for JAFFE. For the 100-Driver dataset, $\alpha = 0.1$ and $\alpha = 0.3$ both result in comparable performance, but for better privacy protection, we select $\alpha = 0.3$.

To explore the impact of optimization on utility models trained with DP-SGD, we compare random feature masking and optimized feature masking. We assess how these methods affect the performance of utility models when training with DP-SGD at $\epsilon$ values that correspond to $ASR^M \approx 0.54$ (marked by the black dotted line in Figure~\ref{fig: Differential Privacy: privacy}). Figure~\ref{fig:Differential Privacy: acc} and Table~\ref{tab:summary} demonstrate that applying optimized feature masking to datasets results in significantly improved $\epsilon$ and performance of utility models when $ASR^M \approx 0.54$, compared to solely using DP-SGD. We conclude that utility models trained on datasets processed by optimized feature masking offer comparable privacy protection while allowing for much larger $\epsilon$ values during DP-SGD training. This increased $\epsilon$ enables the utility models to achieve a better privacy-utility tradeoff compared to models trained on the original datasets or those using random feature omission.

\begin{table}[htbp]
\centering
\renewcommand{\arraystretch}{1.2} 
\setlength{\tabcolsep}{12pt} 
\begin{tabular}{|c|c|c|c|c|}
\hline
\multirow{2}{*}{\textbf{Dataset}} & \multicolumn{2}{c|}{\textbf{Orig. DP-SGD}} & \multicolumn{2}{c|}{\textbf{FM+DP-SGD}} \\
\cline{2-5}
 & $\epsilon$ & Acc & $\epsilon$ & Acc \\
\hline
JAFFE      & 1.73 & 0.1824 & 6.87 & 0.335 \\
\hline
RaFD       & 1.65 & 0.1717 & 6.54 & 0.291 \\
\hline
TFEID      & 1.78 & 0.1762 & 4.33 & 0.287 \\
\hline
100-Driver & 2.13 & 0.5731 & 7.14 & 0.637 \\
\hline
\end{tabular}
\caption{$\epsilon$ and Accuracy of utility models when $ASR^M \approx 0.54$ (the black dotted line in Figure~\ref{fig: Differential Privacy: privacy_LIME}).}
\label{tab:summary}
\end{table}

\subsection{Result Analysis: Impact of Explainer}
In the previous experiments, we have used SHAP to implement the feature masking algorithm. It’s also interesting to understand whether the method is explainer-specific. In this section, we replace SHAP with LIME and reproduce the results. As shown in Figure~\ref{fig: Differential Privacy: privacy_LIME} and Figure~\ref{fig:combined_LIME}, LIME does not change the patterns much for both privacy protection and utility preservation on the facial expression datasets. However, on the 100-Driver dataset, the optimized FM curve overlaps with the random FM curve, even with a smaller $\alpha=0.1$ (Figure~\ref{fig:100Driver(LIME)}), as opposed to a larger $\alpha=0.3$ for SHAP (Figure~\ref{fig:100Driver(SHAP)}). This suggests that LIME-based optimized FM may not perform as well as SHAP on more complex model architectures. We also observe that models using LIME exhibit larger standard deviations in accuracy (Figure~\ref{fig:Differential Privacy: acc_Lime}) compared to Figure~\ref{fig:Differential Privacy: acc}. Figure~\ref{fig:stability} looks into the stability of SHAP and LIME when applied with optimized FM at $\epsilon=7$. SHAP consistently provides better stability across all datasets. We attribute this to LIME being a localized interpretation method \cite{ribeiro2016why}, which is more appropriate for simpler models where localized interpretability suffices. Moreover, LIME's reliance on random sampling may introduce additional variance compared to SHAP.

In conclusion, when deploying optimized FM in practice, SHAP appears to be the more suitable backbone explainer than LIME.

\subsection{Result Analysis: Time Cost}

\begin{figure}[h]
    \centering
    \begin{subfigure}[b]{0.48\textwidth}
        \centering
        \begin{tikzpicture}[scale=0.8]
            \begin{axis}[
                ybar,
                bar width=10pt,
                width=\textwidth,  
                height=0.4\textwidth,
                symbolic x coords={JAFFE, RaFD, TFEID, 100-Driver, IN},
                xtick=data,
                ylabel={Time (s)},
                ymin=0, ymax=20,
                error bars/.cd,
                enlarge x limits=0.15,
                ymajorgrids=true,
                grid style=dashed
            ]
            \addplot+[
                ybar,
                color=blue,
                fill=blue!30,
                error bars/.cd,
                y dir=both,
                y explicit
            ] coordinates {
                (JAFFE, 2.6) +- (0,0.04)
                (RaFD, 2.9) +- (0,0.03)
                (TFEID, 2.3) +- (0,0.05)
                (100-Driver, 6.36) +- (0,0.11)
                (IN, 7.17) +- (0,0.46)
            };
            \end{axis}
        \end{tikzpicture}
        \caption{Explainer (SHAP) time cost per image}
        \label{explanation_time}
    \end{subfigure}
    \hfill
    \begin{subfigure}[b]{0.48\textwidth}
        \centering
        \begin{tikzpicture}[scale=0.8]
            \begin{axis}[
                ybar,
                bar width=10pt,
                width=\textwidth,  
                height=0.4\textwidth,
                symbolic x coords={JAFFE, RaFD, TFEID, 100-Driver},
                xtick=data,
                ylabel={Time (min)},
                ymin=0, ymax=7,
                error bars/.cd,
                enlarge x limits=0.15,
                ymajorgrids=true,
                grid style=dashed
            ]
            \addplot+[
                ybar,
                color=red,
                fill=red!30,
                error bars/.cd,
                y dir=both,
                y explicit
            ] coordinates {
                (JAFFE, 3.63) +- (0,0.21)
                (RaFD, 3.48) +- (0,0.15)
                (TFEID, 3.21) +- (0,0.21)
                (100-Driver, 5.25) +- (0,1.24)
            };
            \end{axis}
        \end{tikzpicture}
        \caption{Optimized FM generation time cost per class}
        \label{optimization_time}
    \end{subfigure}
    \caption{Time cost of sensitivity generation by SHAP and optimizing feature masks across datasets.}
    \label{time_cost}
\end{figure}
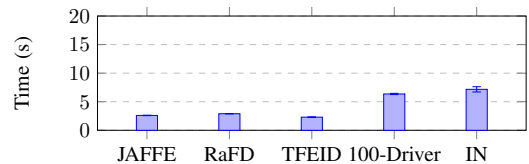
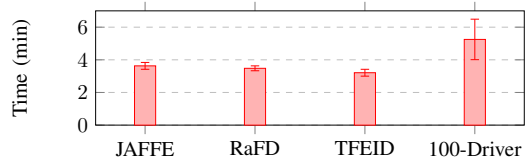

In this section, we evaluate the time cost of the optimized feature masking (FM) method. This approach involves two main steps: (1) generating feature privacy and utility sensitivity levels for each image with an explainer, and (2) performing class-wise optimization of the feature masks. Figure~\ref{time_cost} presents the average time cost for three facial expression datasets using the ResNet-18 model and for the 100-Driver dataset using the ResNet-101 model with SHAP as the explainer. To assess the potential time cost for larger, real-world datasets, we also include the time cost for the ImageNet-1K (IN) dataset on the ResNet-101 model.
Although the per-image time cost remains in the seconds range (Figure~\ref{explanation_time}), thanks to SHAP’s batch-processing capability, the total time for large datasets like 100-Driver and ImageNet-1K can accumulate to hours or even days. Methods like sampling can be applied to reduce the first-stage cost. In contrast, the optimization of feature masks takes only a few minutes for the facial expression datasets, 100-Driver, and ImageNet-1K (Figure~\ref{optimization_time}). 

\subsection{Discussion}
\label{sec:limit}
We have experimented with a variety of datasets to observe the data- and model-specific characteristics of $ASR^M$. Among these datasets, we have used facial expression and 100-driver datasets for identity-related feature masking experiments. For datasets with no clear identity elements, like those used in self-driving car technology featuring pedestrians, buildings, and address plates, defining identity tasks can be trickier, probably tightly related to the analysis of application-specific sensitivity.  Nevertheless, our work has established a framework to understand the inherent sensitivity of the dataset and guide the setting of privacy parameters for differential privacy methods. For a specific application and dataset, one can always try different identity-related tasks and evaluate them within our framework. 

\section{Conclusion}
\label{sec:conclusion}
The parameter setting of differentially private machine learning is detached from specific applications, datasets, and models, which is considered a unique feature. However, to better understand the tradeoff between privacy and utility, e.g., finding a justifiable relaxed $\epsilon$ setting, we have to look into more data- and model-specific auxiliary measures. Recent studies on likelihood-ratio-based membership inference attacks, LiRA, have given us special tools to tackle this challenging problem. We have shown that the LiRA-based attacking success rate $ASR^M$ provides a well-justified data- and model-specific measure for evaluating practical privacy risks for models trained with or without DP-SGD. We explore the factors affecting this attacking success rate to identify a better way to set the theoretical differential privacy budget. With the proposed optimized feature masking methods, we demonstrated in experiments that models trained on datasets with masked features and relaxed $\epsilon$ settings can still achieve equivalent practical privacy protection (i.e., similar $ASR^M$ levels) compared to original DP-SGD. Meanwhile, the optimization method preserves the utility-critical features and, thus, better model quality. Our approach offers a promising framework for fine-tuning the privacy budget setting in terms of specific data and models to achieve better privacy and utility balances.
\bibliographystyle{IEEEtran}
\bibliography{ref}
\end{document}